\documentclass[11pt,a4paper]{article}
\usepackage[hyperref]{emnlp-ijcnlp-2019}
\usepackage{times}
\usepackage{latexsym}
\usepackage[utf8]{inputenc}  
\usepackage{url}
\usepackage{graphbox}
\aclfinalcopy 
\usepackage{xcolor}
\usepackage{examples}
\usepackage[normalem]{ulem} 
\usepackage{soul}
\newcommand{\featName}[1]{\textsc{#1}}
\newcommand{\litlex}[1]{\uwave{#1}}  
\newcommand{\lex}[1]{\textbf{#1}}  
\newcommand{\ile}[1]{\textsl{#1}} 
\newcommand{\lit}[1]{`\textsl{#1}'} 
\newcommand{\idio}[1]{`#1'}  
\newcommand{\exlit}[2]{\ile{#1}~\lit{#2}} 
\newcommand{\exlitidio}[3]{\ile{#1}~\lit{#2}$\Rightarrow$\idio{#3}} 

\usepackage{amsmath}
\usepackage{tikz}
\newcommand*\circled[1]{\tikz[baseline=(char.base)]{
            \node[shape=circle,draw,inner sep=2pt] (char) {#1};}}
\usepackage{hhline} 
\usepackage{hyperref}
\usepackage{todonotes}

\usepackage{verbatim}
\def\checkmark{\tikz\fill[scale=0.2](0,.35) -- (.25,0) -- (1,.7) -- (.25,.15) -- cycle;}

\usepackage{multirow}
\usepackage{array,multirow,makecell} 
\setcellgapes{1pt} 
\makegapedcells 
\newcolumntype{C}[1]{>{\centering\arraybackslash }b{#1}} 
\newcommand{\STAB}[1]{\begin{tabular}{@{}c@{}}#1\end{tabular}}

\title{To Be or Not To Be a Verbal Multiword Expression:\\A Quest for Discriminating Features}

\author{Caroline Pasquer \\
  University of Tours\\ France \\
  {\tt caroline.pasquer@gmail.com} 
    \\\And
   Agata Savary, Jean-Yves Antoine\\
  University of Tours\\ France \\
   {\tt first.last@univ-tours.fr}  
  \\\AND
   Carlos Ramisch \\
  Aix Marseille Univ, Université de Toulon,\\CNRS, LIS, Marseille, France \\
  {\tt carlos.ramisch@lis-lab.fr} 
  \\\And
   Nicolas Labroche, Arnaud Giacometti\\
  University of Tours\\ France \\
   {\tt first.last@univ-tours.fr}  
     \\
}
\date{}

\begin{document}
\maketitle
\begin{abstract}
%
Automatic identification of mutiword expressions (MWEs) is a pre-requisite for semantically-oriented downstream applications. This task is challenging because MWEs, especially verbal ones (VMWEs), exhibit surface variability. However, this variability is usually more restricted than in regular (non-VMWE) constructions, which leads to various variability profiles. We use this fact to determine the optimal set of features which could be used in a supervised classification setting to solve a subproblem of VMWE identification:  the identification of occurrences of previously seen VMWEs. Surprisingly, a simple custom frequency-based feature selection method proves more efficient than other standard methods such as Chi-squared test, information gain or decision trees. 
An SVM classifier using the optimal set of only 6 features outperforms the best systems from a recent shared task on the French seen data.


\end{abstract}

\section{Introduction}
\label{sec:intro}
Multiword expressions (MWEs) are word combinations with idiosyncratic characteristics regarding for instance morphology, syntax or semantics \cite{baldwin-kim:2010:handbook}. One of their most emblematic properties is semantic non-compositionality: the meaning of the whole expression cannot be straightforwardly deduced from the meaning of its components, as in \exlit{to \lex{cut corners}}{to do an incomplete job}\footnote{Henceforth, the lexicalized components of a MWE, i.e.~those always realized by the same lexemes, appear in bold.}. Due to this property, as well as to their pervasiveness \cite{Jackendoff1997}, MWEs constitute a major challenge for semantically-oriented downstream applications, such as machine translation, information retrieval or sentiment analysis. A prerequisite for an appropriate handling of MWEs is their automatic identification.

MWE identification aims at automatic location of MWEs in running text. This task is very challenging, as signaled by \citet{Constantetal17}, and further confirmed by the PARSEME Shared Task on automatic identification of verbal MWEs \cite{p:ramisch-etAl:2018:lawmwecxg}. One of the main difficulties stems from the variability of MWEs, especially verbal ones (VMWEs). Namely, even if a MWE has previously been observed (in a training corpus or in a lexicon), it can re-appear in morphosyntactically diverse forms, where components vary inflectionally, their order is inverted, discontinuities occur and syntactic relations change between lexicalized components, as in examples (\ref{ex:cut-corners-1}--\ref{ex:cut-corners-2}). 
\vspace{-0.1em}
\begin{examples}
\small
\item\label{ex:cut-corners-1}
Some companies were \lex{cutting corners}$_{\textsc{obj}}$ to save costs.

\vspace{-0.3em}
\item\label{ex:cut-corners-2} The field would look uneven if \lex{corners}$_{\textsc{subj}}$ were \lex{cut}.
\end{examples}
\vspace{-0.2em}
However, assuming unrestricted variability is not a good strategy either, since it may lead to literal or coincidental occurrences of MWEs' components, as in (\ref{ex:cut-corners-3}) and (\ref{ex:cut-corners-4}).\footnote{Henceforth, literal and coincidental occurrences are highlighted with wavy underlining.} 
\begin{examples}
\small
\item\label{ex:cut-corners-3} Start with \litlex{cutting} one \litlex{corner} of the disinfectant bag.
\vspace{-1em}
\item\label{ex:cut-corners-4} If you \litlex{cut} along this line, you'll get an acute \litlex{corner}.
\end{examples}
\vspace{-0.2em}
To tackle these challenges, we focus on a subproblem of MWE identification (henceforth referred to as the Task), namely the identification of previously seen VMWEs. 

We deal with French, which exhibits a particularly rich verbal inflection, and whose VMWE-annotated corpus is one of the richest PARSEME Shared Task benchmark datasets. However, the proposed methods are language-independent in that they rely on universal properties of VMWEs and feature sets.
Our aim is to determine the optimal set of features which would allow us to automatically discriminate between VMWEs and non-VMWEs in a supervised classification framework. 
To this aim, we first study the linguistic properties of VMWEs which may serve as a basis for defining the initial set of features (Sec.~\ref{sec:vmwes}). Then, we describe the corpora used for training and evaluation (Sec.~\ref{sec:corpus}). Further, we address the selection of the positive/negative candidates from the corpus (Sec.~\ref{sec:extraction}). We discuss 
the successive phases of feature selection (Sec.~\ref{sec:feature-select}): the definition of the initial set of features; the ranking of these features by their relevance to the task; and establishing the optimal number of best-ranked features by off-the-shelf binary classifiers.
Then, the results are discussed (Sec.~\ref{sec:results}) and we interpret some of the selected features from a linguistic point of view (Sec.~\ref{sec:lingRelevance}). Finally, we conclude and suggest directions for future work (Sec.~\ref{sec:conclusion}).

\section{Linguistic properties of VMWEs}
\label{sec:vmwes}
The major observation underlying our work is that VMWEs are mostly morphosyntactically regular at the level of tokens (individual occurrences) but idiosyncratic at the level of types (sets of surface realizations of the same expression). For instance, considering the VMWE type \ile{to \lex{cut corners}} instantiated by the VMWE tokens in (\ref{ex:cut-corners-1}) and (\ref{ex:cut-corners-2}), we find no "surface" (i.e.~non-semantic) hints that distinguish them from regular English verb-object constructions (e.g.~\ile{cut branches}). 
They respect standard grammar rules (the object follows the verb; in passive, the verb occurs in participle, etc.). Unlike named entities, they use no capitalisation and contain no trigger words. 
However, number inflection of the noun is prohibited by this VMWE, i.e.~using the noun in singular leads to the loss of the idiomatic meaning, as shown in example (\ref{ex:cut-corners-3}). This is in contrast to regular verb-object constructions (\ile{cut a branch}), in which noun inflection does not significantly change the overall meaning.

Such a restricted variability of MWEs is one of their fundamental properties, as argued in linguistic work \cite{GastonGross88,Nunberg1994,Tutin2016,Sheinfux2018}. Namely, a given syntactic structure in a given language comes with a set of various morphosyntactic variations considered grammatical (e.g.~English transitive verb constructions admit passivisation, noun inflection, pronominalisation, etc.). But a MWE of the same structure usually allows only a proper subset of these variants, which we will call its \emph{variability profile}. These profiles are numerous and hard to predict, e.g.~while \ile{\lex{cut corners}} admits passivisation but prohibits number inflection, \exlit{\lex{build bridges}}{establish links} accepts both of these variants, and \exlit{\lex{take place}}{happen} none of them. 

Studying variability profiles in corpora is hindered by another fundamental  property of VMWEs -- their Zipfian distribution: few VMWE types occur frequently in corpora, and there is a long tail of VMWE types occurring rarely. 

The third property we are interested in is the fact that some VMWE types do share common "surface" properties, e.g.~most light-verb constructions have no lexicalized determiners (\ile{\lex{take} a/many/several/no \lex{break(s)}})
and contain frequent light verbs (e.g.~\ile{\lex{take} a \lex{walk}/\lex{break}/\lex{advantage}/etc.}). 
These shared properties, however, are rarely semantically motivated: VMWEs exhibit a strong lexical inflexibility, i.e.~replacing a lexicalized component with a semantically close word (a synonym, hyperonym, etc.) usually results in the loss of the idiomatic reading, in example (\ref{ex:cut-corners-5}), as opposed to (\ref{ex:cut-corners-1}--\ref{ex:cut-corners-2}).

\vspace{-0.3em}
\begin{examples}
\small
\item\label{ex:cut-corners-5} The field would look uneven if \underline{borders} were \underline{reduced}.
\end{examples}
\vspace{-0.3em}

In this work, we hypothesize that: (i) va\-ria\-bi\-li\-ty profiles of VMWEs, materialized by morphosyntactic features in annotated corpora, can help correctly identify occurrences of previously seen VMWEs, (ii) for rarely occurring VMWEs, their resemblance with other, more frequent, VMWEs can also help solve the Task. The two hypotheses give rise to \emph{relative} and \emph{absolute} features, described in Sec.~\ref{sec:initial-features}.

\section{Corpus}
\label{sec:corpus}
In 2018, edition 1.1 of the PARSEME Shared Task, henceforth PST \citep{p:ramisch-etAl:2018:lawmwecxg}, took place, with a goal of boosting the development of identifiers for both seen and unseen VMWEs. To this end, a corpus of verbal MWEs was manually annotated and openly published for 19 languages.\footnote{{\scriptsize \url{http://hdl.handle.net/11372/LRT-2842}}} 
We use the French part of this corpus, and we refer to its particular subcorpora as TrainST and TestST, which reflects their function in PST.

In our experiments, TrainST is used: (i) for training and testing during feature selection (Sec.~\ref{sec:extraction}--\ref{sec:features}), (ii) for training during the evaluation of the final system against the PST results  (Sec.~\ref{sec:benchmark}), and prior to the extraction of candidates for manual evaluation on a large external corpus (Sec.~\ref{sec:manual-eval}). TestST is only used as a testing benchmark in the PST evaluation setting (Sec.~\ref{sec:benchmark}).\footnote{In the PST corpus, there is also a small development corpus for French, which we do not use.}

For manual evaluation (Sec.~\ref{sec:manual-eval}), we also use a large corpus collected from Wikipedia and by web\-crawling for the CoNLL 2017 Shared Task.\footnote{{\scriptsize \url{http://hdl.handle.net/11234/1-1989}}}

It is automatically segmented, tokenized and morpho-syntactically annotated \cite{zeman2018conll} but it does not contain VMWE annotations.

Tab.~\ref{tab:corpus} shows the statistics of the 3 corpora, in terms of the number of sentences, tokens, all annotated VMWEs and those annotated VMWEs which also occur in TrainST (the Task is limited to them). 

All corpora comply with the CoNLL-U format\footnote{{\scriptsize \url{http://universaldependencies.org/format.html}}}, i.e. contain, for each surface form, its lemma, POS, morphological features and syntactic dependencies.
 
VMWEs are manually annotated and categorized into: verbal idioms (VID: \ile{\lex{cut corners}}), light-verb constructions (LVC: \ile{\lex{take} a \lex{walk}}), inherently reflexive verbs (IRV: \exlitidio{\lex{s'apercevoir}}{to perceive oneself}{to realize}) and  multi-verb constructions (MVC: \ile{to \lex{make do}}). Thus, contrary to other works which focus on verb-noun VMWEs \cite{DBLP:journals/coling/FazlyCS09}, we handle various syntactic profiles. Note, however, that the Task only addresses VMWE identification and abstracts away from categorisation.

\begin{table}[t!]
\footnotesize
\begin{center}
\setlength{\tabcolsep}{0.2mm}
\begin{tabular}{|l|r|r|r|rr|}
\hline \textbf{Corpus} & \textbf{\# Sentences} &\textbf{\# Tokens} & \textbf{\# VMWEs} & \multicolumn{2}{c|}{\textbf{\# Seen}} \\ \hline
TrainST & 17,225& 432,389 & 4,550 & 4,550 & {\scriptsize 100\%}\\
TestST &1,606 & 39,489 & 498 & 251 & {\scriptsize 50\%}\\
CoNLL  & 306,431,406& 5,242,235,570 & n/a & n/a & \\
\hline
\end{tabular}
\end{center}
\caption{Statistics about corpora. \emph{Seen} refers to those VMWEs which also appear in TrainST.}
\label{tab:corpus}
\end{table}

\section{Candidate extraction}
\label{sec:extraction}
In order to test the hypotheses put forward in Sec.~\ref{sec:vmwes}, we 
propose a supervised classification method for identifying previously seen VMWEs. 

In the first step, we design a procedure, henceforth called \texttt{ExtractCands}, to extract VMWE candidates. It will be employed in: (i) training, to choose positive and negative VMWE examples, further used for feature selection,  (ii) testing, for pre-identifying VMWE candidates to be fed to the classifier. 
For every VMWE $e$ attested in the training corpus, \texttt{ExtractCands} extracts each co-occurrence $c$ of $e$'s lexicalized components, provided that the following conditions are fulfilled. 

First, the sets of lemmas and parts-of-speech of $e$'s and $c$'s components are identical. For instance, for $e$ in (\ref{ex:take-measure-1}), the candidate in (\ref{ex:take-measure-2}) will be extracted but not the one in (\ref{ex:take-measure-3}), due to the different POS of \exlit{mesure}{measure}.

\begin{figure*}[tb!]
\centering
\setlength{\tabcolsep}{0.1mm}
\begin{tabular}{cc}
{\scriptsize (a)}\includegraphics[scale=0.7,align=c]{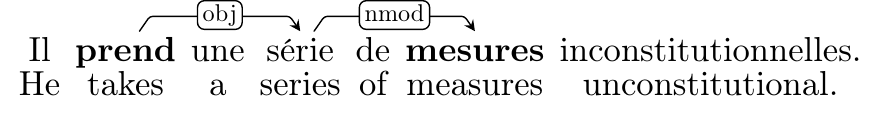} &
{\scriptsize (b)}\includegraphics[scale=0.7,align=c]{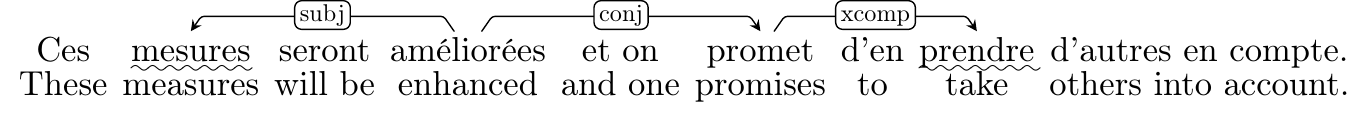}
\end{tabular}
\caption{VMWE candidates with discontinuous dependency chains: (a) extracted, (b) non-extracted.}
\label{fig:take-measure-4}
\end{figure*}

\vspace{-0.5em}
\begin{examples}
\small
\item\label{ex:take-measure-1} Il \lex{prend}$_{\textsc{verb}}$ plusieurs$_{\textsc{det}}$ \lex{measures}$_{\textsc{noun}}$. 'He takes several measures.'
\vspace{-0.8em}
\item\label{ex:take-measure-2} Je \lex{prends}$_{\textsc{verb}}$ une$_{\textsc{det}}$ \lex{mesure}$_{\textsc{noun}}$ unconstitutionnnelle. 'I take an unconstitutional measure.'
\vspace{-0.8em}
\item\label{ex:take-measure-3} Il \litlex{prend}$_{\textsc{verb}}$ une règle et \uwave{mesure}$_{\textsc{verb}}$ la longueur. 'He takes a ruler and measures the length.'
\end{examples}
\vspace{-0.5em}
Second, if $c$ contains two components or if there is only one verb and one noun, they must be either directly connected in the syntactic dependency tree or separated by only one element. This condition is fulfilled (\ref{ex:take-measure-2}) and in case of complex determiners as in Fig.~\ref{fig:take-measure-4}(a), but not in coincidental occurrences like in Fig.~\ref{fig:take-measure-4}(b). No dependency constraints were defined for candidates with more than two components due to their lower frequency and to computational constraints.\footnote{Searching for existing, direct or indirect, dependencies between pairs taken from up to 7 components would be  time-consuming.} 

Third, based on $c$'s discontinuities seen in TrainST, we restrict the number of external elements that can be inserted between $c$'s components, to avoid large numbers of spurious candidates stemming from frequent lemmas (e.g.~determiners or pronouns), as in (\ref{ex:il-faut-2}).
\vspace{0.2em}
\begin{examples}
\small
\vspace{-0.6em}
\item\label{ex:il-faut-2}Il \litlex{faut} rappeler que, jusqu'en 1983, \litlex{il} n'y avait pas \ldots\\
One must remind that, until 1983, there was no \ldots
\end{examples}
\vspace{-0.4em}

Finally, since we wish to test the variability profile hypothesis (Sec.~\ref{sec:vmwes}), we retain only those VMWEs whose number of occurrences is high enough to be representative of their variability. However, this frequency threshold cannot be too high, otherwise the size of the annotated data would dramatically drop. For a reasonable trade-off between these two factors, we select only those candidates whose attested VMWEs appear at least twice in the training corpus (3582 occurrences i.e.~78\% of TrainST). 

When \texttt{ExtractCands} is used for training, the extracted candidates are marked as positive, if they are manually annotated as VMWEs in TrainST, and negative otherwise. Tab.~\ref{tab:corpusExtr} shows the results of the candidate extraction in the corpora from Tab.~\ref{tab:corpus}. As expected, the method is tuned for a high recall and a reasonable precision. The latter factor should grow due to classification, based on a carefully selected set of features, which is addressed in the following section.

\begin{table}[t!]
\footnotesize
\begin{center}
\setlength{\tabcolsep}{0.7mm}
\begin{tabular}{|l|r|r|r|c|c|}
\hline \multirow{2}{*}{\textbf{Corpus}} & \multicolumn{3}{c|}{\textbf{Extracted candidates}} & \multirow{2}{*}{\textbf{P}} & \multirow{2}{*}{\textbf{R}} \\\cline{2-4}
  & \textbf{All} & \textbf{Positive} &\textbf{Negative} & & \\\hline
  TrainST & 4,596 & 3,582 \tiny(78\%)&1,014 \tiny(22\%)& 0.78 & 0.98 \\
TestST & 368 & 210 \tiny(57\%)&158 \tiny(43\%)&0.57 & 1.00 \\ \hline
CoNLL  & 32,789,815 & n/a &n/a& n/a & n/a \\\hline
\end{tabular}
\end{center}
\caption{Results of candidate extraction, based on  attested VMWEs seen at least twice in TrainST.
}
\label{tab:corpusExtr}
\end{table}
\section{Feature selection}
\label{sec:features}
\label{sec:feature-select}
In supervised classification, the VMWE candidates to classify are to be represented by sets of features. The choice of such features is crucial for the quality of the classification outcome and its adequacy for the tested hypotheses. This section describes the initial set of features and the selection of those which are the most relevant to the Task.

\subsection{Initial set of features}
\label{sec:initial-features}

As discussed in Sec.~\ref{sec:vmwes}, we hypothesize that the variability profiles of VMWEs can help correctly identify occurrences of previously seen VMWEs. Representing these profiles  straightforwardly is not always easy (e.g.~for passivization or pronominalization), especially in a language-independent way. But, we assume that they can be approximated by simpler properties directly encoded in the CoNLL-U files (Sec.~\ref{sec:corpus}). Thus, our starting point is a set of features related to morphology, syntax and insertions. Morphological features mostly refer to verb or noun inflection. Syntactic features relate to dependency relations outgoing from a component (e.g.~subject vs.~object) or to the fact that the components are syntactically (in)directly connected. Insertion-bound features account for the number of words inserted between the lexicalized components, e.g.~2 insertions in (\ref{ex:take-measure-insert-1}), and to their POS sequence, e.g.~\textsc{det adj} in (\ref{ex:take-measure-insert-1}). 
Insertions and syntactic dependencies may convey similar information, e.g.~in (\ref{ex:take-measure-insert-1}) the adjective is accounted for both as an insertion and as a modifier of the noun. But they may also be complementary, as in (\ref{ex:take-measure-insert-2}), since syntactic dependencies capture modifiers not included between the lexicalized components.

\vspace{-0.2em}
\begin{examples}
\small
\item\label{ex:take-measure-insert-1}  Il \lex{prend} d'$_{\textsc{det}}$ importantes$_{\textsc{adj}}$ \lex{mesures} 'He \lex{takes} $\emptyset$ $_{\textsc{det}}$ important \lex{measures}'

\vspace{-0.8em}
\item\label{ex:take-measure-insert-2} \exlit{Il \lex{prend} des$_{\textsc{det}}$ \lex{mesures} aussi (voire de plus en plus) importantes$_{\textsc{adj}}$}{He \lex{takes} as (or even more and more) important \lex{measures}}

\end{examples}
\vspace{-0.5em}

Recall (Sec.~\ref{sec:vmwes}) that the variability profile is defined on the level of VMWE types rather than tokens. This is why, following \citet{PasquerST2018-W18-4932}, we put strong impact on \emph{relative} features. Namely, given a VMWE candidate $c$, and its reference VMWE token together with all attested VMWE tokens of the same type ($\{e_1, e_2,\ldots,e_n\}$), we compare the properties of $c$ to those of $e_i$ ($1 \leq i \leq n$). Relative features always take binary values. Thus, the \featName{REL\_insertSeq} feature is true if the POS sequence of the words inserted between $c$'s components is the same as in any $e_i$, as in (\ref{ex:take-measure-2}) vs.~(\ref{ex:take-measure-1}), and false otherwise, as in (\ref{ex:take-measure-insert-2}) vs.~(\ref{ex:take-measure-insert-1}). Also, \featName{REL\_syntacticDependencies\_VERB} is true if the set of dependency relations outgoing from the verb in $c$ is the same as in any $e_i$, and false otherwise, etc.

Recall that VMWEs have a Zipfian distribution, thus relative features may be unreliable for many VMWE types occurring rarely. Thus, we also use \emph{absolute} features. They are similar to the relative ones, but they are categorical rather than binary. For instance, the value of  \featName{ABS\_insertSeq} is the POS sequence of the inserted words, e.g.~\textsc{det adj} in (\ref{ex:take-measure-insert-1}). Absolute features also include the VMWE set of lemmas, e.g.~$\{mesure, prendre\}$, and their categorization (e.g.~LVC).

Note finally that no word embeddings are used in the initial set of features. This is justified by lexical inflexibility of VMWEs: a new VMWE can rarely be discovered by semantic similarity of its components to those of attested VMWEs. This observation is confirmed by the PST, where best VMWE identifiers, even those using word embeddings, never exceed F-measure of 0.28 for unseen VMWEs. In our opinion, this very weak generalisation power should be approached by systematically coupling VMWE identification with their automatic discovery, which our perspective for future work.

\subsection{Usefulness of automatic feature selection}
\label{sec:usefulness}

The initial set of relative and absolute features defined in the previous section is huge: with the UD tagset used in our corpora, we get 18,000 possible features in total. If we only keep those features which are activated at least once in the corpus, this number is reduced to about 800. 
The question is then how to select those features which are the most discriminating for the Task. 

Most systems from PST use features which are known to be linguistically relevant, and which are relatively simple to calculate \cite{Waszczuk-W18-4931,shoma-DBLP:journals/corr/abs-1809-03056,GBD-DBLP:conf/acllaw/BorosB18,stodden2018trapacc,moreau2018crf,moreau2018crf,berk2018deep,ehren2018mumpitz,PasquerST2018-W18-4932,zampieri2018veyn}.
\footnote{A summary of this state of the art is offered in the Appendix B.}

Another approach was to feed all/most potentially relevant features to the classifier and let it determine the appropriate weights for those features \cite{PasquerST2018-W18-4932}. In this work, we also rely on domain knowledge to define the initial set of features, but we use automatic feature selection methods. The motivation is threefold. First, a lower number of accurately chosen features should lead to classification of a higher quality and shorter training time. Second, we might discover features whose relevance to the Task has not been linguistically established. Third, automatic procedure is language-independent, as soon as the initial set of features is generic enough.

The feature selection process runs in two stages: (i) feature ranking, (ii) choice of the optimal number of the best-ranked features.

\subsection{Feature ranking}
\label{sec:feature-ranking}

We experimented with 4 feature ranking methods described below. All but the first require annotated data and are applied to the TrainST corpus.

\noindent \textbf{\texttt{FREQ}} This is a custom frequency-based method. We apply the \texttt{ExtractCands} from Sec.~\ref{sec:extraction} to the CoNLL corpus, which yields 2.6 million candidates.\footnote{Their classes are unknown, since the CoNLL corpus in not annotated for VMWEs.} Then, we select only those feature-value pairs which appear in at least one candidate, and we rank them by decreasing frequency. Finally, we strip the values and only keep the features.\footnote{After stripping, features may have several occurrences, corresponding to different values. We only keep the occurrence with the highest rank.} 

\noindent \textbf{\texttt{CHI2}} The Chi-squared test \cite{kumbhar2016survey} looks at candidates having both the given feature-value pair $fv$ and the given class $cl$. It calculates the difference between their observed and expected frequencies, under the hypothesis that $fv$ and $cl$ are independent. We use the Pearson's version of this test, and the feature-value ranking corresponds to decreasing chi-squared value. We finally strip the values, as in \texttt{FREQ}.

\noindent \textbf{\texttt{GAIN}} The information gain \cite{kumbhar2016survey} method partitions the set of observations according to different values of a given feature $f$, calculates the weighted sum of the entropies of these partitions, and compares it to the entropy of the whole set. If the entropy strongly drops after the partition, the information gain is high, i.e.~$f$ is strongly discriminating. The feature ranking corresponds to the decreasing value of the gain.

\noindent \textbf{\texttt{FOREST}} The Random Forest algorithm \cite{Breiman2001}, combines many decision trees (here: 10 trees with no depth limit) into a single model by a majority or weighted vote. 
For the construction of the tree, the quality of each split is measured by the Gini impurity. Features are then ranked by their relevance when determining splits. 

\subsection{Tuning the number of features}
\label{sec:tuning}

Each of the methods from the previous section ranks the initial set of features according to their estimated relevance to the Task. We now need to decide how many best-ranked features from each ranking should be selected. We tune this parameter with 3 standard supervised classification methods --  Naive Bayes, linear SVM and decision trees -- using their off-the-shelf implementations.\footnote{NLTK \cite{Loper02nltk:the} for Naive Bayes, and Scikit-learn \cite{scikit-learn} for SVM and decision trees.}

For each classifier $C$, and for each feature ranking list $R$ from Sec.~\ref{sec:feature-ranking}, we proceed by a greedy method : (i) we select the $i$ best-ranked features from $R$, (ii) we train and test $C$, with the $i$ selected features, in a 10-fold unstratified cross-validation setting with a 90\%-10\% split of the TrainST corpus, where the candidates for both training and testing are those extracted by \texttt{ExtractCands}, (iii) we calculate the mean F-measure $F_{mean}$ for the 10 folds, (iv) we repeat steps (i--iii) for every $i$ from 1 to $length(R)$ and we select the value of $i$ for which $F_{mean}$ is the highest.

\section{Results}
\label{sec:results}
In this section we show the results of the feature selection process described above. Then, we evaluate the obtained set of features by two methods: by the comparison of the best-performing classifier-selection pair to benchmark results, and by a manual evaluation of the same pair on an external large corpus.

\subsection{Optimal feature sets}
\label{sec:optimal-sets}

Tab.~\ref{tab:resultsFeatSelect} shows the best results (i.e.~the results for the optimal number of best-ranked features) of the feature selection experiments described in Sec.~\ref{sec:features}, for the 4 feature ranking methods and the 3 classifiers. $P$, $R$ and $F$ are the mean scores from the 10 folds, and $\sigma$ is the variance of the 10 F-scores.\footnote{The low value of variance indicates that the simple 90\%-10\% corpus split method is sufficiently reliable. If variance were higher, stratified validation would be more appropriate (i.e.~the corpus would have to be split in such a way that the proportion of data belonging to each class is the same in each fold as in the whole corpus).} 

Surprisingly, \texttt{FREQ}, which is a custom method, initially conceived to only remove irrelevant features from the initial set, achieves identical or better results
than standard feature selection methods, with F = 0.936 for the linear SVM.

\begin{table}[t!]
\footnotesize
\begin{center}
\begin{tabular}{|@{}l@{}|@{}c@{}|@{}l@{}|@{}c@{}|@{}c@{}|@{}c@{}|@{}c@{}|}
 \hline
\textbf{\textbf{FeatSelect}}&\textbf{Feature set  }&\texttt{\textbf{Classif}} &\textbf{P} &\textbf{R} & \textbf{F} &$\sigma$\\ \hline
& \circled{1}&NB&0.881&0.965&0.921&0.012\\ \cline{2-7} 
\texttt{FREQ} & \circled{2}&SVM&0.896&0.980&\textbf{0.936}&0.015\\
&  &Dec.~Tree&0.913&0.934&0.923&0.009\\\hline
                & &NB&0.818&0.975&0.889&0.011\\
\texttt{CHI2}   & \circled{3}&SVM&0.826&0.973&0.893&0.013\\
                 & &Dec.~Tree&0.833&0.964&0.894&0.012\\\hline
 & &NB&0.826&0.974&0.894&0.013\\
\texttt{GAIN}   &\circled{4} &SVM&0.897&0.980&\textbf{0.936}&0.015\\
              & &Dec.~Tree&0.875&0.962&0.916&0.020\\\hline
                                &&NB&0.787&0.975&0.871&0.016\\
\texttt{FOREST}  & \circled{5}&SVM&0.787&0.975&0.871&0.016\\
               & &Dec.~Tree&0.787&0.975&0.871&0.016\\\hline

\end{tabular}
\end{center}
\caption{\label{tab:resultsFeatSelect}Best mean performance on 10\% of TrainST (in 10 cross-fold validation). The feature sets are detailed in Tab.~\ref{tab:setFeatures}.}
\end{table}

The optimal feature sets corresponding to these results are described in Tab.~\ref{tab:setFeatures}. 
There is a notable consistence in the optimal sets selected by each method. For \texttt{CHI2}, \texttt{GAIN} and \texttt{FOREST}, each of the 3 classifiers selects always the same set (resp.~\circled{3}, \circled{4}, \circled{5}). For \texttt{FREQ}, the two sets (\circled{1} and \circled{2}) vary by one feature only.
However, the sets are quite diverse from one selection method to another, which proves the complementarity of the methods.

The optimal number of features varies from 4 to 8.  
As expected, relative features are more often selected;  \texttt{CHI2} and \texttt{FOREST} select even exclusively relative features. This is consistent with our variability-profile hypothesis (Sec.~\ref{sec:vmwes}). 
One feature (\featName{REL\_insertSeq}) is selected by all methods, and there are 6 features selected only once.

\begin{table}[t!]
\small
\begin{tabular}{|@{}l@{}|l@{}|l@{}|l@{}|l@{}|l@{}|@{}c@{}|}
\hline
\textbf{ABSolute and RELative}&\multicolumn{5}{c|}{\textbf{Feature sets in Tab.~\ref{tab:resultsFeatSelect}}}&Total\\
\textbf{features}

&\tiny\circled{1}&\tiny\circled{2}&\tiny\circled{3} &\tiny\circled{4} &\tiny\circled{5} &\\ \hline
\featName{ABS\_LemmaSet}&\checkmark&\checkmark&&\checkmark&&3\\
(e.g.~\{\lex{\ile{prendre}},\ile{\lex{décision}}\}&&&&&&\\\hline
\featName{ABS\_VMWEcat}&\checkmark&\checkmark&&&&2\\
(e.g.~LVC)&&&&&&\\\hline
\featName{ABS\_insertSeq} &\checkmark&\checkmark&&\checkmark&&\textbf{3}\\
(e.g.~DET-ADJ) &&&&&&\\\hline
\featName{ABS\_lemma\_VERB} &&\checkmark&&&&1\\
 (e.g.~'\ile{\lex{prendre}}')&&&&&&\\\hline\hline
\featName{REL\_insertSeq} &\checkmark&\checkmark&\checkmark&\checkmark&\checkmark&\textbf{5}\\
\featName{REL\_insert\_ADP} &&&\checkmark&\checkmark&\checkmark&3\\
\featName{REL\_insert\_CCONJ}&&&&&\checkmark&1\\
\featName{REL\_insert\_DET}&&&\checkmark&\checkmark&&2\\
\featName{REL\_insert\_NOUN}&&&\checkmark&\checkmark&\checkmark&3\\
\featName{REL\_insert\_PUNCT}&&&\checkmark&&&1\\
\featName{REL\_insert\_VERB}&&&\checkmark&&&1\\
\featName{REL\_0to5insertions}&&&\checkmark&\checkmark&&2\\
\featName{REL\_lemma\_VERB}&&\checkmark&&&&1\\
\featName{REL\_SyntacticDistance}&&&\checkmark&&&1\\\hline
Number of features&4&6&8&7&4&\\\hline
\end{tabular}
\caption{\label{tab:setFeatures} Optimal feature sets mentioned in Tab.~\ref{tab:resultsFeatSelect}} 
\end{table}

\subsection{Benchmark evaluation}
\label{sec:benchmark}

It is interesting to evaluate the quality of the feature set selection by 
a direct comparison to the 17 systems having participated in the PST campaign (cf. Sec.~\ref{sec:corpus}). PST defined evaluation measures which are specific to phenomena constituting known challenges in VMWE identification. One of them is the seen-in-train score, i.e.~the P/R/F calculated only for those VMWEs in TestST which were previously seen in TrainST.  In PST, a VMWE from TestST is considered seen if a VMWE with the same (multi-)set of lemmas is annotated at least once in TrainST. 

We compare our optimal $\circled{2}$+SVM classifier with the PST results for French\footnote{{\scriptsize \url{https://gitlab.com/parseme/sharedtask-data/tree/master/1.1/system-results}}} as follows: (i) SVM is trained on the whole TrainST corpus (with \texttt{ExtractCands} and feature set $\circled{2}$), (ii) we apply it to the candidates extracted with \texttt{ExtractCands} from TestST, (iii) those VMWEs from TestST which were either omitted by \texttt{ExtractCands} or wrongly classified simply count as false negatives. This obviously lowers the results as compared to Tab.~\ref{tab:resultsFeatSelect}, where only the VMWEs seen at least twice are considered.

As shown in Tab.~\ref{tab:resultsCompar}, the F-measure of $\circled{2}$+SVM indeed drops by over 11 points in the PST setting (to 0.8207). However, this is much higher than $F=0.7003$ obtained by varIDE \cite{PasquerST2018-W18-4932} with a quite similar method, except that, there, only Naive Bayes with no feature selection is used. This highlights the fact that focusing on a small set of relevant features helps the classification. What is more, our F-measure is slightly higher than the best registered F-score (0.8172) by TRAVERSAL  \cite{Waszczuk-W18-4931}, where logistic regression is applied to sequences of nodes in dependency trees. Additionally, the handling of the variant-of-train VMWEs (i.e.~those VMWEs in TestST which appear with a different surface form than in TrainST) is better with our method than with TRAVERSAL (F = 0.7317 vs.~0.7123). We also get better seen-in-train F-scores than the second and third best systems, based on neural networks: GBD-NER-resplit \cite{GBD-DBLP:conf/acllaw/BorosB18}, which only uses the provided PST corpora, and SHOMA \cite{shoma-DBLP:journals/corr/abs-1809-03056}, which also employs Wikipedia word embeddings.

When $\circled{2}$+SVM is evaluated on all VMWEs from the French TestST, not only the seen ones, the F-score decreases to 0.55, which is still relatively close to 0.56 obtained by the best system (TRAVERSAL), even if unseen VWMES are totally beyond the scope of our work. 

\begin{table}[t!]
\footnotesize
\begin{center}
 \begin{tabular}{|l|@{}c|@{}c|@{}c|}
\hline 
\multicolumn{4}{|c|}{Evaluation on 10\% TrainST (seen twice VMWEs)}\\\hline \hline
Method&P&R&F\\\hline 
Feature set \circled{2} + SVM&0.896&0.980&\textbf{0.936}\\\hline  
\hline\hline
\multicolumn{4}{|c|}{Evaluation on TestST (all seen VMWEs)}\\\hline \hline
Feature set \circled{2} + SVM &0.8207&0.8207&\textbf{0.8207}\\\hline
TRAVERSAL&0.8879&0.7570&0.8172\\
GBD-NER-resplit&0.8842&0.6693&0.7619\\
SHOMA&0.9167&0.6574&0.7657\\
varIDE&0.5682&0.9124&0.7003\\\hline\hline
\multicolumn{4}{|c|}{Evaluation on WebSample}\\\hline \hline
Feature set \circled{2} + SVM &0.873&0.977&\textbf{0.922}\\\hline
\end{tabular}
\end{center}
\caption{\label{tab:resultsCompar} Evaluation results in the PST benchmark setting, and with a manually annotated external corpus} 
\end{table}

\subsection{Manual evaluation on an external corpus}
\label{sec:manual-eval}

It is also interesting to evaluate the quality of the selected feature set on an external corpus, independent of the one used for the selection. To this end we constructed a representative sample of the CoNLL corpus, called WebSample, that could be manually checked. Namely, WebSample 
should respect the distribution observed in TrainST, as far as frequencies of VMWEs and of their categories are concerned. To this aim, we first checked the proportion of the 4 categories (VIDs, LVCs, IRVs and MVCs) in the VMWEs seen at least twice in TrainST. Then, we selected 90 VMWE types
\footnote{cf. Appendix A.}
in such a way that these proportions are respected (except for MVCs, which are very rare in TrainST) and that VMWEs types in each category have high, low and median frequency (with a balanced distribution). For MVCs, we selected all 4 VMWE types appearing in TrainST. Finally, the 90 VMWE types selected in TrainST were given as the reference set to \texttt{ExtractCands}, which was then applied to the CoNLL corpus. The resulting set of candidates was trimmed so that equal numbers of candidates come from the Wikipedia and the webcrawling part of the CoNLL corpus.

Tab.~\ref{tab:sample} shows the outcome of this procedure. The resulting set of 4645 candidates (from 90 types), together with their occurrence contexts, was then manually annotated by 4 experts. 
27 occurrences (0.6\%) were eliminated due to incorrect sentences, errors (POS, lemma), non-French words or insufficient context. This low rejection rate confirms the good overall quality of the CoNLL corpus. 
The 4618 remaining candidates were manually labelled as positive (68.7\%) and negative.

\begin{table}[t!]
\footnotesize
\begin{center}
\begin{tabular}{|l@{}|@{}c|@{}c|@{}c|@{}c|@{}c|@{}c@{}|}
\hline
&\multicolumn{3}{c}{Candidates in TrainST} & \multicolumn{3}{|c|}{Candidates in WebSample} \\ \cline{2-7} 
\textbf{Cat.} & \textbf{\# types} &\textbf{\# occ.} & \textbf{\%}  & \textbf{\# types} &\textbf{\# occ.} & \textbf{\%} \\\hline
VID	&180&		1404&	39,2&	30	&1900&40.9	\\
IRV	&144	&	1136&	31,7&	29	&1508	&32.5\\
LVC	&263&		1024&	28,6&	27	&1213&26.1	\\
MVC	&4	&	18&	0,5	&4	&24	&0.5\\\hline
Total	&591&		3582&	100&	90&4645	&100\\
\hline
\end{tabular}
\end{center}
\caption{\label{tab:sample} WebSample w.r.t.~TrainST (VMWEs seen at least twice)}
\vspace{-1em}
\end{table}

The WebSample corpus was then used as the testing corpus for our $\circled{2}$+SVM classifier. As shown in Tab.~\ref{tab:sampleResultPercat}, we obtained the overall F-measure of 0.922, which is comparable to the 0.936 achieved on TrainST and highlights the good repeatability of the method whatever the corpus: our system suffers neither from overfitting nor from sensitivity to the data source. As to per-category results disregarding the MVCs (whose number is negligible), LVCs are the hardest to classify ($F=0.903$), probably due to their high variability, as illustrated in examples (\ref{ex:result1}--\ref{ex:result5}).

\begin{table}[t!]
\footnotesize
\begin{center}
\begin{tabular}{|l|c|c|c|}
\hline
\textbf{VMWE category}&\textbf{P}&\textbf{R}&\textbf{F}\\\hline
VID&0.917&0.961&0.939\\
IRV&0.873&0.985&0.926\\
LVC&0.830&0.990&0.903\\
MVC&0.750&0.750&0.750\\\hline
All&0.873&0.977&0.922\\
\hline
\end{tabular}
\end{center}
\caption{\label{tab:sampleResultPercat} SVM results on WebSample per category}
\end{table}

\section{Linguistic relevance of the selected features}
\label{sec:lingRelevance}
The selected features prove relevant to the phenomenon of VMWE variability. 
First, as outlined in Tab.~\ref{tab:setFeatures}, the inserted POS sequences like \textsc{det adj} in (\ref{ex:result3}) always stand at the top of feature rankings (as a REL and/or an ABS feature), whereas this feature is generally not taken into account in other works.
Insertions can, indeed, model many relevant phenomena such as the determiner flexibility in (\ref{ex:result1}) vs.~(\ref{ex:result2}), allowed modifiers as the adjective in (\ref{ex:result3}), passivization (\ref{ex:result4}) or relativization (\ref{ex:result5}). 
Inserted POSes including \textsc{verb}, \textsc{noun} or \textsc{punct}uation appear as the most relevant, maybe because they may suggest a non-VMWE as in \ile{I \litlex{build}  viaducts}$_{\textsc{noun}}$, \ile{you construct}$_{\textsc{verb}}$ \ile{\litlex{bridges}}.
\begin{examples}
\small
\item\label{ex:result1} \exlit{Je \lex{prends} des$_{\textsc{det}}$ \lex{décisions}}{I make decisions}
\vspace{-0.8em}
\item\label{ex:result2} \exlit{Je \lex{prends} deux$_{\textsc{num}}$ \lex{décisions}}{I make two decisions}
\vspace{-0.8em}
\item\label{ex:result3} \exlit{Je \lex{prends} une$_{\textsc{det}}$ grande$_{\textsc{adj}}$ \lex{décision}}{I make a great decision}
\vspace{-0.8em}
\item\label{ex:result4} \exlit{Ma \lex{décision} est$_{\textsc{aux}}$ \lex{prise}}{My decision is made}
\vspace{-0.8em}
\item\label{ex:result5} \exlit{C'est la \lex{décision} que$_{\textsc{pron}}$ je$_{\textsc{pron}}$ \lex{prends}}{It is the decision I make}
\vspace{-0.8em}
\end{examples}
Note that 3 of the 6 selected features are lexical: they relate to the lemmas of the VMWE components. Without these features, our F-score decreases by 4 points. 
This confirms the strength of the lexical inflexibility phenomenon (Sec.~\ref{sec:vmwes}) and the hardness of generalization over unseen data, observed in PST.\footnote{The best average PST results limited to unseen data are below $F=0.29$, even for systems using neural networks and word embeddings  \cite{p:ramisch-etAl:2018:lawmwecxg}.} 

Some features, considered relevant to VMWE variability, do not appear as selected in Tab.~\ref{tab:setFeatures}. E.g.~many VMWEs prohibit the modification of the noun (\ile{to \litlex{cut} \underline{acute} \litlex{corners}}) or its inflection for number (\ile{to \litlex{cut} a \litlex{corner}}). Given that such variations are often tolerated by LVCs but not by VIDs, the VID candidates may not have been sufficiently numerous to acquire such  knowledge about VID constraints.

\section{Conclusions and future work}
\label{sec:conclusion}
We presented a methodology for selecting a relevant set of features which can be used in a supervised classification framework to solve the task of identifying occurrences of previously seen VMWEs. It is based on defining an initial large set of linguistically-motivated features, and then applying 4 feature selection methods, together with 3 classifiers, to select the optimal sets of features.

The results show that relative features (those reflecting the variability profiles of VMWEs) dominate in the optimal selections. Most of the selected features are based on the properties of external insertions. Such features are never or rarely explicitly considered in previous works.  

In sum, our system's main contributions are: (i) its performances, on par, for French, with the best systems of a recent shared task, 
(ii) its ability to highlight linguistic properties of VMWEs by a novel combination of features, (iii) its interpretability and classification efficiency due to only 6 features, 
(iv) its generalization power confirmed by an evaluation on an external large corpus.

The fact that 50\% of the optimal features are lemma-based stresses the lexical nature of the MWE phenomenon, which implies the particularly acute hardness of generalising over unseen data. 
We believe that this hardness cannot be overcome by more powerful features (e.g. stemming from distributional semantics), but should rather be handled by coupling
MWE identification with unsupervised MWE discovery. In this way, large parts of the unseen data can be transformed into seen data, and their corpus occurrences can complement the manually annotated training corpus. We will address such coupling in future work.

Other perspectives include a detailed error analysis as well as finding strategies for VMWEs seen once, which were disregarded here. 
The relevant features likely differ among different VMWE classes, since these classes have different syntactic properties. Therefore, it should be useful to perform the feature selection separately for each class.
Finally, we could evaluate the genericity of our method by evaluating it on other languages of the PARSEME Shared Task. The feature selection across languages or language families might, indeed, reveal universal characteristics of VMWEs. 

\bibliography{biblio}
\bibliographystyle{acl_natbib}
\appendix
\label{sec:appendixA}
\section{Appendix: List of the verbal multiword expressions in WebSample}
\label{sec:appendixB}

\begin{table}[!htbp]
\footnotesize
\begin{center}
\begin{tabular}{|@{}c@{}|@{}c@{}|@{}c@{}|@{}c@{}|@{}c@{}|}
 \hline
	Frequency&	IRV	&	LVC	&	VID		\\
	in TRAIN	&		&		&				\\\hline
	&s'étendre&faire apparition&il s'agir\\
	&se voir&avoir droit&il convenir\\
	&se trouver&jouer match&mettre fin\\
	&se retrouver&poser question&avoir lieu\\
	&s'élever&faire appel&il y avoir\\
High	&se situer&avoir besoin&faire partie\\
	&s'engager&rendre hommage&il falloir\\
	&se produire&avoir chance&tenir compte\\
	&se dérouler&jouer rôle&faire l'objet\\
	&se rendre&&prendre part\\\hline
	&s'efforcer&conclure accord&prendre au piège\\
&se servir&mener opération&en finir\\
&se baser&porter choix&avoir affaire\\
&s'agir&rendre service&venir à bout\\
Median&s'envoler&disputer épreuve&ne pas payer de mine\\
&se lever&avoir perception&vouloir dire\\
&s'adonner&livrer bataille&s'en aller\\
&s'interroger&présenter signe&faire écho\\
&s'élancer&donner concert&mettre sur pied\\
&&&mettre en lumière\\\hline
	&s'acquitter&faire traduction&tourner mal\\
	&s'engouffrer&avoir ennui&couper du monde\\
	&se ressentir&prendre sanction&être au rendez-vous\\
Low	&se donner&faire entrée&tenir responsable\\
	&se partager&inscrire but&faire les frais\\
	&s'empresser&effectuer tournée&être l'occasion\\
	&se renseigner&dresser horoscope&mettre sur la table\\
	&se signaler&signer victoire&prendre le dessus\\
	&s'investir&réserver accueil&tel être le cas\\
&se promener&&faire le plein\\\hline
\end{tabular}
\end{center}
\caption{\label{tab:MWEsinSample} Verbal multiword expression types (for IRV, LVC and VID categories) in WebSample (mentioned in Table 6).}
\end{table}

\begin{table}[!htbp]
\footnotesize
\begin{center}
\begin{tabular}{|c|}
 \hline
	MVC	\\\hline
		entendre parler \\
		faire remarquer \\
		faire savoir\\
		laisser tomber\\\hline
\end{tabular}
\end{center}
\caption{\label{tab:MWEsinSampleMVC} Verbal multiword expression types (for MVC category) in WebSample (mentioned in Table 6).}
\end{table}

\onecolumn
\section*{}
\label{sec:appendixB}
\section{Appendix: Features used for VMWE identification in the state-of-the-art works}
\label{sec:appendixC}
\begin{table*}[ht!]
\footnotesize
\begin{center}
\begin{tabular}{|@{}l@{}|@{}l@{}|@{}C{1cm}@{}|@{}C{0.9cm}@{}|@{}C{1cm}@{}|@{}C{0.9cm}@{}|@{}C{1cm}@{}|@{}C{0.9cm}@{}|@{}C{1cm}@{}|@{}C{0.9cm}@{}|@{}C{1cm}@{}|@{}C{0.9cm}@{}||@{}c@{}|@{}c@{}|}
\hline

			\multicolumn{2}{|c|}{}&	\multicolumn{10}{c||}{Systems of the PARSEME Shared Task}	&		\multicolumn{2}{c|}{Our system}	
				\\\cline{3-14}
		\multicolumn{2}{|c|}{}	&		&	&	 	&		&		&		&		&		&		&	&Initial  	&Auto. 	\\
				\multicolumn{2}{|c|}{ABSolute (categorical) and}
				 	&	(a)	&	(b)	&	(c)	&	(d)	&	(e)	&	(f)	&	(g)	&	(h)	&	(i)	&	(j)&set of	&selected 					\\
				\multicolumn{2}{|c|}{RELative (binary) features}&		&	&	 	&		&		&		&		&		&		&	&	features	&	features	\\\hline
\multirow{15}{*}{\STAB{\rotatebox[origin=c]{90}{ABS}}}&VMWE category (e.g.~LVC) &	yes	&		&		&	yes	&	yes	&		&		&		&	yes	&	yes	&	yes	&	\textbf{yes}	\\\cline{2-14}
	&	VMWE surface form	&	yes	&	yes	&	yes	&	yes	&		&		&	yes	&		&		&		&	no	&	no	\\
	&	VMWE whole set of lemmas 	&		&		&		&		&		&		&		&		&	yes	&		&	yes	&	\textbf{yes}	\\
&	VMWE separate lemmas 	&	yes	&	yes	&	yes	&	yes	&	yes	&	yes	&		&	yes	&	yes	&	yes	&	yes	&	\textbf{yes}	\\
	&		VMWE POS	&	yes	&	yes	&	yes	&	yes	&		&	yes	&	yes	&	yes	&	yes	&		&	yes	&	no	\\
&	VMWE length	&		&		&		&	yes	&		&		&		&		&		&		&	no	&no		\\\cline{2-14}
	&\textit{VMWE morphosyntactic tags}	&		&		&	yes	&	yes	&	yes	&	yes	&		&		&	yes	&		&	yes	&	no	\\\cline{2-14}
&	\textit{VMWE syntactic dependencies}	&		&		&		&		&		&		&		&		&		&		&		&		\\
&	\hspace{0.3cm}Connection	&	yes	&		&		&		&	yes	&		&		&		&	yes	&		&	yes	&	no	\\
&	\hspace{0.3cm}Quasi-connection	&		&		&		&		&		&		&		&		&	yes	&		&	yes	&	no	\\
&	\hspace{0.3cm}Labels	&	yes	&		&		&	yes	&	yes	&		&	yes	&	yes	&	yes	&		&	yes	&	no	\\\cline{2-14}
&	\textit{Insertions}:	&		&		&		&		&		&		&		&		&		&		&		&		\\
&\hspace{0.3cm}Sequences of POS 	&		&		&		&		&		&		&		&		&	yes	&		&	yes	&	\textbf{yes}	\\
&	\hspace{0.3cm}Number	&		&		&		&		&		&		&		&		&	yes	&		&	yes	&	no	\\\hline
\multirow{10}{*}{\STAB{\rotatebox[origin=c]{90}{REL}}}
&	VMWE separate lemmas 	&		&		&		&		&		&		&		&		&	yes	&		&	yes	&	\textbf{yes}	\\
	&		VMWE POS	&		&		&		&		&		&		&		&		&	yes	&		&	yes	&	no	\\\cline{2-14}
	&\textit{VMWE morphosyntactic tags}	&		&		&		&		&		&		&		&		&	yes	&		&	yes	&	no	\\\cline{2-14}
&	\textit{VMWE syntactic dependencies}	&		&		&		&		&		&		&		&		&		&		&		&		\\
&	\hspace{0.3cm}Connection	&		&		&		&		&		&		&		&		&	yes	&		&	yes	&	no	\\
&	\hspace{0.3cm}Quasi-connection	&		&		&		&		&		&		&		&		&	yes	&		&	yes	&	no	\\
&	\hspace{0.3cm}Labels	&		&		&		&		&		&		&		&		&	yes	&		&	yes	&	no	\\\cline{2-14}
&	\textit{Insertions}:	&		&		&		&		&		&		&		&		&		&		&		&		\\
&\hspace{0.3cm}Sequences of POS 	&		&		&		&		&		&		&		&		&	yes	&		&	yes	&	\textbf{yes}	\\
&	\hspace{0.3cm}Number	&		&		&		&		&		&		&		&		&	yes	&		&	yes	&	no	\\\hline
	\multicolumn{2}{|c|}{External word embeddings}	&		&	yes	&		&		&		&		&	yes	&		&		&		&		no	&no	\\\hline\hline
\multicolumn{2}{|c|}{Seen-in-train F-score (French)}	&	\textbf{0.8172} 	&	0.7657	&	0.7619	&	0.7506	&	0.7371	&	0.7290	&	0.7286	&	0.7084	&	0.7003	&	0.6224	&		&	\textbf{0.8207}	\\
\hline
\end{tabular}
\end{center}
\caption{\label{tab:featuresSOA-ident} Relevant features for VMWE identification and performances for the previously seen VMWEs for our system vs.~the systems in the PARSEME Shared Task \cite{p:ramisch-etAl:2018:lawmwecxg}: 	
(a) \citet{Waszczuk-W18-4931}, (b) \citet{shoma-DBLP:journals/corr/abs-1809-03056}, (c) \citet{GBD-DBLP:conf/acllaw/BorosB18}, (d) \citet{stodden2018trapacc},	(e) \citet{moreau2018crf} (CRF-DepTree-categs system), (f) \citet{moreau2018crf} (CRF-Seq-nocategs system), (g) \citet{berk2018deep}, (h) \citet{ehren2018mumpitz}, (i) \citet{PasquerST2018-W18-4932}, (j) \citet{zampieri2018veyn}}
\vspace{-1em}
\end{table*}
\end{document}